\documentclass[conference]{IEEEtran}
\IEEEoverridecommandlockouts

\usepackage{cite}
\usepackage{amsmath,amssymb,amsfonts}
\usepackage{algorithmic}
\usepackage{graphicx}
\usepackage{siunitx}
\usepackage{textcomp}
\usepackage{xcolor}
\usepackage{booktabs}
\usepackage{pgfplots}
\pgfplotsset{compat=1.18}
\def\BibTeX{{\rm B\kern-.05em{\sc i\kern-.025em b}\kern-.08em
    T\kern-.1667em\lower.7ex\hbox{E}\kern-.125emX}}

\IEEEoverridecommandlockouts
\begin{document}
\IEEEpubid{
  \makebox[\columnwidth]{
    Accepted to IEEE BSN 2025. \copyright~2025 IEEE \hfill
  }
  \hspace{\columnsep}\makebox[\columnwidth]{ }
}

\title{Domain-Specific Constitutional AI: Enhancing Safety in LLM-Powered Mental Health Chatbots\\

}

\author{
Chenhan Lyu$^1$, Yutong Song$^1$, Pengfei Zhang$^1$, Amir M. Rahmani$^1$ \\
\IEEEauthorblockA{$^1$University of California, Irvine, USA \\
\{clyu4, yutons12, pengfz5, amirr1\}@uci.edu
}
}

\maketitle

\begin{abstract}
Mental health applications have emerged as a critical area in computational health, driven by rising global rates of mental illness, the integration of AI in psychological care, and the need for scalable solutions in underserved communities. These include therapy chatbots, crisis detection, and wellness platforms handling sensitive data, requiring specialized AI safety beyond general safeguards due to emotional vulnerability, risks like misdiagnosis or symptom exacerbation, and precise management of vulnerable states to avoid severe outcomes such as self-harm or loss of trust. Despite AI safety advances, general safeguards inadequately address mental health-specific challenges, including crisis intervention accuracy to avert escalations, therapeutic guideline adherence to prevent misinformation, scale limitations in resource-constrained settings, and adaptation to nuanced dialogues where generics may introduce biases or miss distress signals. We introduce an approach to apply Constitutional AI training with domain-specific mental health principles for safe, domain-adapted CAI systems in computational mental health applications.
\end{abstract}


\section{Introduction}

As developers continue to create increasingly advanced large language models for everyday applications, ensuring their safety and alignment with human preferences remains a paramount concern~\cite{wang2024comprehensive, anwar2024foundational, hubinger2024sleeper}. Techniques such as reinforcement learning from human feedback (RLHF) and Constitutional AI (CAI) have emerged as key methods for creating helpful and harmless assistants, with CAI enabling self-critique and revision guided by explicit principles~\cite{bai2022constitutional, bai2022training, kundu2023specific}. The integration of Large Language Models (LLMs) into mental health applications represents both tremendous opportunity and significant risk in computational health systems, encompassing therapy chatbots, crisis detection algorithms, and personalized wellness platforms that handle sensitive data and vulnerable user states~\cite{guo2024large, lawrence2024opportunities, hua2025scoping}.

Although LLMs could predict problems, intervene and engage in therapeutic dialogues, unaligned models risk producing harmful results, prompting the need for specialized guardrails, evaluation tools, and regulatory guidelines tailored to the nuanced challenges of mental health~\cite{guo2024large, dong2024building}.  While prior research has examined the trade-offs between specific and general principles in CAI training, demonstrating that the general principles can effectively mitigate the majority of harmful responses without requiring exhaustive lists of targeted rules~\cite{kundu2023specific}, and has investigated CAI's applicability to smaller language models through self-critique mechanisms~\cite{menke2025how}, as well as frameworks for systematically crafting and evaluating AI constitutions to enhance alignment~\cite{kyrychenko2025c3ai}, a notable gap persists in the literature. To date, no research has compared constitutional principles explicitly derived from domain-specific mental health guidelines or baselines that forgo constitutional training altogether. This gap is especially pronounced in computational mental health applications, where AI systems must balance therapeutic efficacy with ethical imperatives, potentially leading to unintended risks like inappropriate advice or exacerbated user distress if principles are not finely tuned to the domain's unique demands.

We introduce an approach for applying CAI training with domain-specific mental health principles to develop safe, domain-adapted CAI systems in computational mental health applications. This approach addresses the unique requirements of mental health AI by integrating explicit, tailored principles that prioritize harmlessness, therapeutic accuracy, and ethical alignment, while enabling scalable deployment. It builds on established CAI methodologies to incorporate self-critique mechanisms guided by mental health-specific guidelines, such as those for crisis detection and personalized interventions, ensuring AI responses remain helpful, honest, and sensitive to user vulnerabilities. We advocate for the broader adoption of CAI in specialized domains like mental health and demonstrate its potential through three key contributions:

\begin{itemize}
\item The design of domain-specific constitutional principles derived from mental health guidelines, enabling fine-grained control over AI behaviors in sensitive scenarios like crisis intervention and therapeutic dialogues.
\item A quantitative evaluation comparing these domain-specific principles against general ethical frameworks and baselines without constitutional training, showing enhancements in safety measures, transparency, and adherence to regulatory standards.
\item An exploration of principled training approaches to empower smaller, resource-efficient models to potentially match or surpass larger baselines, facilitating practical implementation in constrained healthcare environments while promoting equity and accessibility in mental health support.
\end{itemize}

\section{Methods}

\subsection{Constitutional Principle}

We derive and apply domain-specific constitutional principles to guide the CAI training process, with the goal of aligning LLMs to the unique safety and ethical demands of mental health applications. These principles are systematically extracted and adapted from comprehensive guidelines on the opportunities and risks associated with LLMs in mental health contexts~\cite{lawrence2024opportunities, guo2024large, hua2025scoping}. The derivation process involves a multi-step approach: first, identifying core themes such as crisis intervention protocols, therapeutic guideline adherence, bias mitigation in emotional dialogues, and user vulnerability handling; second, translating these themes into explicit, actionable rules; and third, refining the principles through iterative review to ensure they are concise yet comprehensive, facilitating effective self-critique and revision during training.

We also establish multiple variants to explore the impact of principle specificity. These include a baseline with no constitutional training, relying solely on the original model; a variant using vague, general ethical principles drawn from foundational AI safety literature, such as broad directives on "promoting user well-being" or "avoiding harm" without domain tailoring~\cite{bai2022constitutional, kundu2023specific, kyrychenko2025c3ai}; a variant incorporating our specific, mental health-adapted principles; and a larger-scale model benchmark without constitutional training to assess scalability aspects. The training pipeline adapts established CAI techniques~\cite{bai2022training, menke2025how}, where models generate initial responses, critique them against the assigned principles, and revise accordingly in a reinforcement learning-inspired loop.

To illustrate the differences between vague/general and specific constitutional principles, we present a side-by-side comparison in Table~\ref{tab:principles_comparison}, which outlines the four principles incorporated into our design.

\begin{figure}[htbp]
\centering
\includegraphics[width=0.5\textwidth, height=0.25\textheight, keepaspectratio=false]{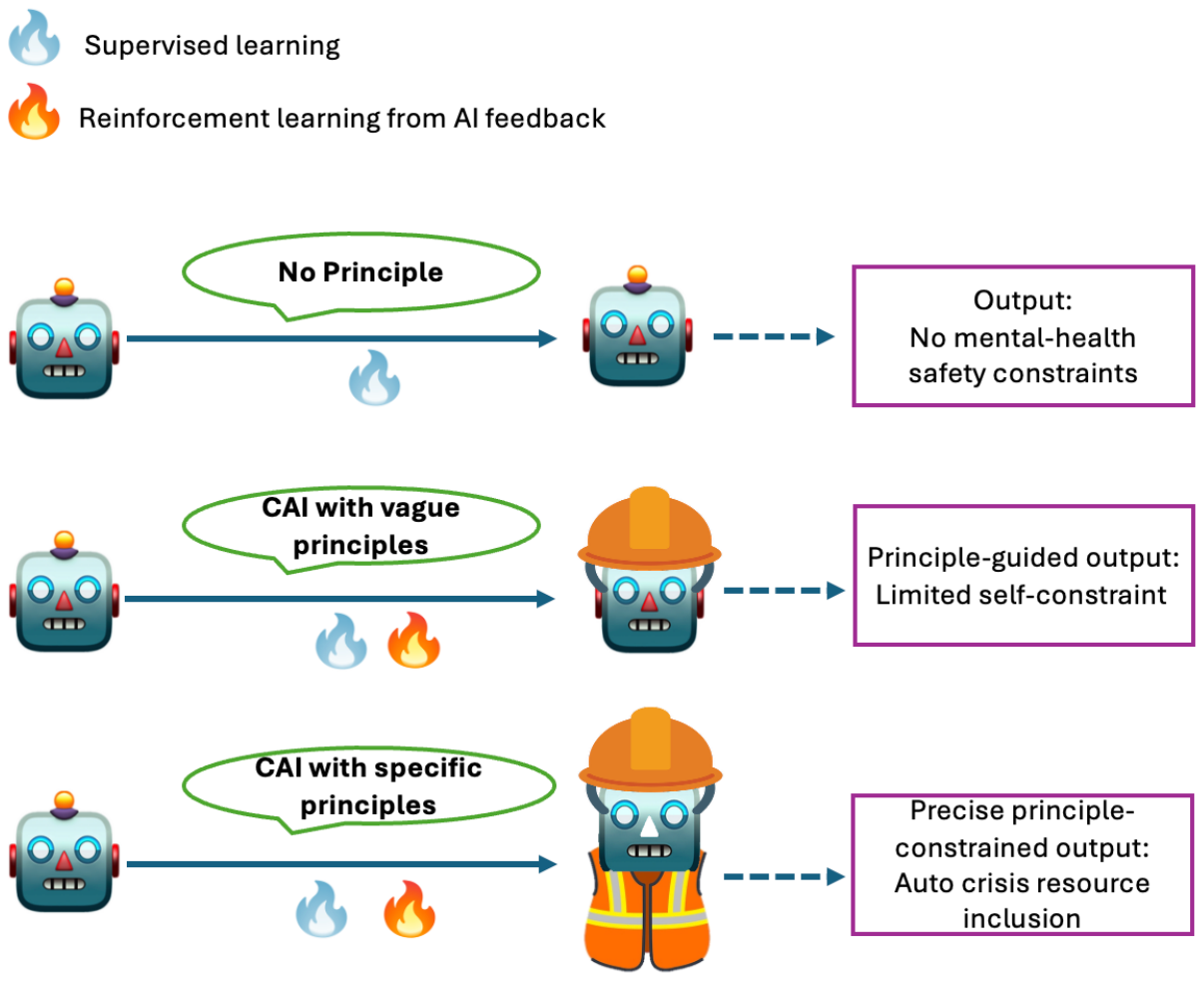}
\caption{Three types of model: model without CAI alignment, model with vague/general principle, model with specific domain related principle}
\label{fig:model}
\end{figure}

\begin{table*}[t]
\centering
\caption{Comparison of Vague/General and Specific Constitutional Principles}
\label{tab:principles_comparison}
\begin{tabular}{p{3cm} p{6cm} p{6cm}}
\toprule
\textbf{Category} & \textbf{Vague/General Principle} & \textbf{Specific Principle} \\
\midrule
Professional Help & Prioritize safety and avoiding harm & Use professional help for serious mental health concerns \\
\midrule
Self-Care Suggestions & Promoting user well-being & Provide evidence-based self-care suggestions \\
\midrule
Language and Tone & Convey understanding & Use empathetic, non-judgmental language \\
\midrule
Crisis Resources & Ensure access to critical resources & Include relevant crisis resources (988 Suicide \& Crisis Lifeline) \\
\bottomrule
\end{tabular}
\end{table*}

\subsection{Constitutional AI Training}
We implemented a four-condition experimental design that includes: (1) a baseline with no additional training, (2) the original model trained with vague/general principles, (3) the original model trained with additional specific derived principles, and (4) a larger model with no constitutional training. We trained two model variants—one with vague/general principles and the other with specific domain-adapted principles—while the baseline and larger model rely on pretrained architectures without further alignment training. The baseline, vague/general, and specific variants utilized the 1B parameter LLaMA 3.2 architecture, while the larger model employed the 3B parameter LLaMA 3.2 architecture, selected to balance computational efficiency with scalability assessments for healthcare deployment~\cite{grattafiori2024llama3herdmodels}.

The CAI training process for the two trained variants (vague/general and specific) follows the established two-phase methodology~\cite{bai2022constitutional, bai2022training}: a supervised fine-tuning (SFT) phase for self-critique and revision, followed by a reinforcement learning from AI feedback (RLAIF) phase for alignment refinement~\cite{wang2024comprehensive, ji2024aligner}. In the SFT phase, the model is prompted to generate initial responses to input queries, critique those responses against the assigned constitutional principles (vague/general or specific, depending on the variant), and produce revised responses that better adhere to the principles. This process leverages chain-of-thought prompting, where the model explicitly reasons about conformance to each principle before revision.

During the RLAIF phase, we generate preference labels tailored to the respective constitutions. Specifically, for each training example, the model samples multiple response pairs. These pairs are then evaluated using AI self-assessment: the model is prompted to compare them head-to-head and assign preference labels based on which response better aligns with the given principles as shown in the table. This process is repeated for both the vague/general and specific variants.

The dataset consists of clinical conversations from MentalChat16K~\cite{MentalChat16K}, sampling 5000 rows, with early stopping to prevent overfitting. We sampled 2 response pairs per example during preference generation, and used standardized prompting templates (e.g., "Critique this response against these principles: [principle text].") to maintain consistency across variants.

\section{Results}
\subsection{Evaluation Framework}
We use the evaluation metric developed by the Institute for Future Health~\cite{park2024building}, which includes five guideline questions with ground truth for mental health chatbot evaluation: Each model was evaluated based on 100 mental health-related queries about common scenarios like depression, anxiety, crises, and general mental wellness. Health experts provided ideal responses as ground truth. Trained evaluators scored responses on a 1-10 scale per guideline using detailed rubrics aligned with clinical best practices. Total scores represented the sum across all five guidelines, with a maximum possible score of 50 per response.

\subsection{Guideline Analysis}
\begin{figure}[htbp]
    \centering
    \includegraphics[width=0.4\textwidth]{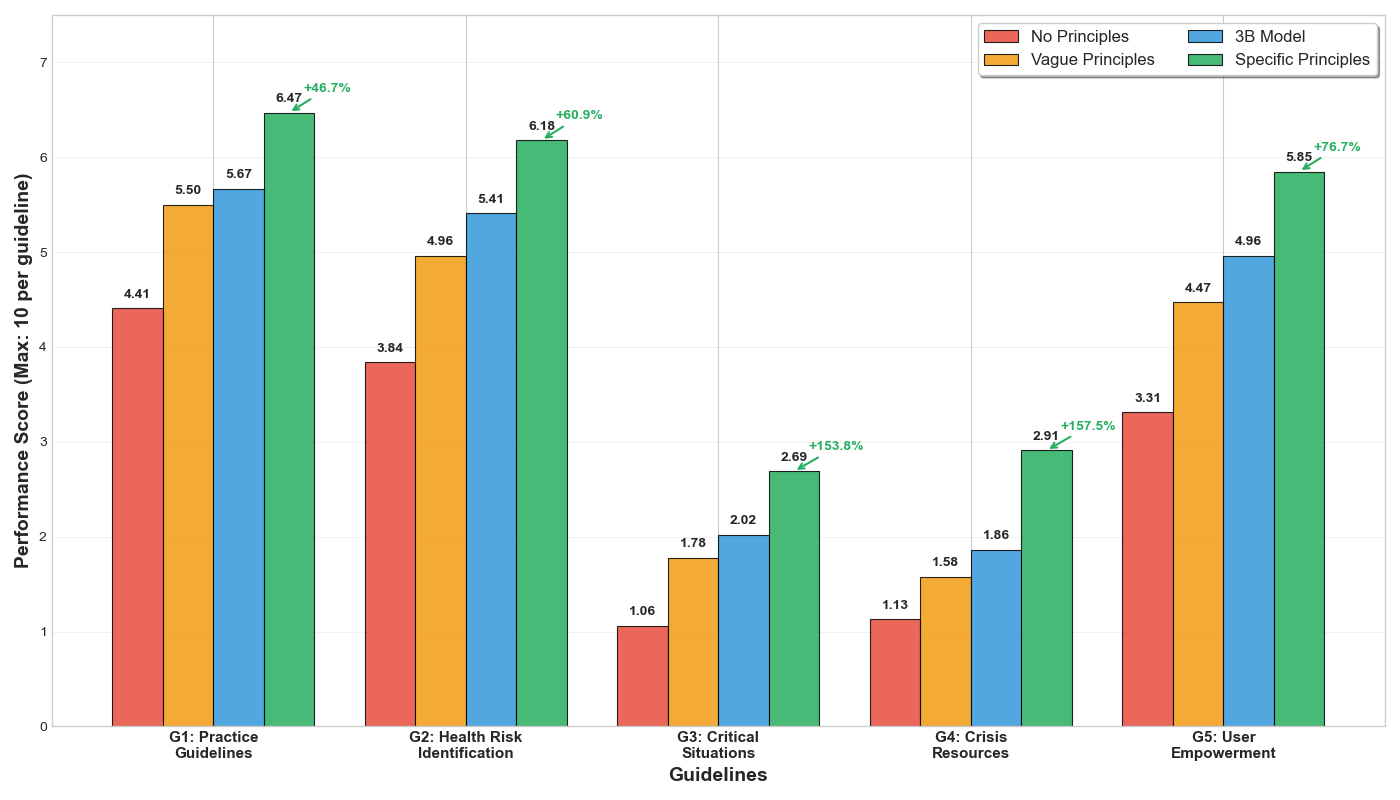}
    \caption{Individual Guideline Performance Analysis}
    \label{fig:figure1}
\end{figure}

Our analysis reveals distinct performance patterns across individual guidelines, as visualized in the bar charts (Figure 2), with the specific-principles model demonstrating the most substantial improvements during the transition from no principles to principled approaches. Guideline 1 (adherence to practice guidelines) shows the strongest absolute gains, improving from 4.41 in the baseline to 6.47 with specific principles (46.7\% increase), with steady progress across conditions: vague/general principles (+24.7\%), 3B model scaling (+28.6\%), and specific principles (+46.7\%), indicating high responsiveness to structured guidance in preventing harmful advice. Guidelines 2 (health risk identification) and 5 (user empowerment) exhibit consistent enhancements, rising from 3.84 to 6.18 (60.9\%) and 3.31 to 5.85 (76.7\%), respectively, with additional boosts of 24.6\% and 30.9\% from vague/general to specific principles, highlighting improved balance in health guidance and professional referrals. Guidelines 3 (consistent response in critical situations) and 4 (resource provision for crisis) display the most pronounced relative improvements despite lower absolute scores, advancing from 1.06 to 2.69 (153.8\%) and 1.13 to 2.91 (157.5\%), underscoring the value of explicit rules for crisis intervention, such as immediate resource provision and help-seeking encouragement in suicidal ideation cases.

While the 3B model provides modest gains over the baseline, smaller models trained with specific principles consistently outperform it—highlighting that principled alignment outweighs scale in critical health interactions. This cross-guideline robustness validates the framework's effectiveness, as explicit rules on medical boundaries, crisis resources, and referrals yield fundamental improvements across contexts—from general adherence to emergency responses—demonstrating CAI's potential for reliable, domain-specific alignment in diverse health domains.

\begin{table*}[t]
\centering
\caption{Model Statistics}
\setlength{\tabcolsep}{8pt} 
\begin{tabular}{l | S[table-format=1.2] S[table-format=1.2] S[table-format=1.2] S[table-format=1.2] S[table-format=1.2] S[table-format=2.2]}
\toprule
\textbf{Model} & \textbf{Guideline1} & \textbf{Guideline2} & \textbf{Guideline3} & \textbf{Guideline4} & \textbf{Guideline5} & \textbf{TotalScore} \\
\midrule
No Principle & 4.41 & 3.84 & 1.06 & 1.13 & 3.31 & \textbf{13.74} \\
Vague/General Principles & 5.50 & 4.96 & 1.78 & 1.58 & 4.47 & \textbf{18.29} \\
No principle (3B) & 5.67 & 5.41 & 2.02 & 1.86 & 4.96 & \textbf{19.92} \\
Specific Principles & \textbf{6.47} & \textbf{6.18} & \textbf{2.69} & \textbf{2.91} & \textbf{5.85} & \textbf{24.08} \\
\bottomrule
\end{tabular}
\end{table*}
\subsection{Efficiency Demonstration}
Previous work has shown the power of small models with CAI~\cite{menke2025how}. We use a more efficient model in the healthcare setting to demonstrate that our method is powerful and efficient. Notably, our 1B parameter model trained with specific principles significantly outperformed a 3B parameter baseline model without constitutional training in preliminary comparisons. This demonstrates that principled training methodologies can overcome scale limitations, a critical finding for resource-constrained healthcare environments where computational efficiency directly impacts deployment feasibility.

The efficiency advantage has important implications for healthcare deployment scenarios, including on-device processing for privacy-sensitive applications, integration with existing hospital IT infrastructure, and accessibility for smaller healthcare institutions with limited computational resources.

\begin{figure}[htbp]
    \centering
    \includegraphics[width=0.27\textwidth]{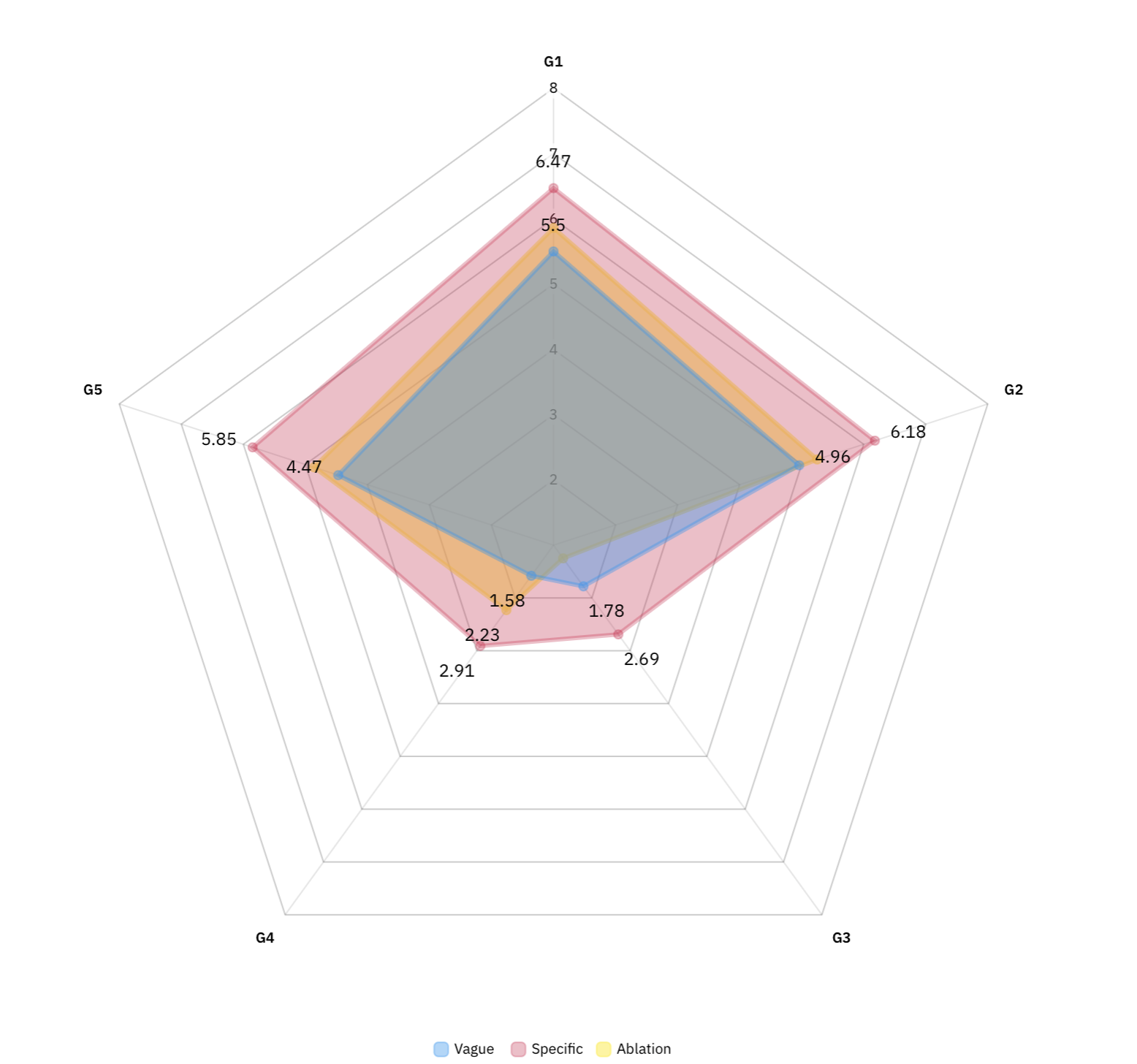}
    \caption{Radar Chart Comparison: Specific Principles Model, Vague Principles Model, and Ablation Model}
    \label{fig:figure2}
\end{figure}
\subsection{Ablation Study}
To dissect the contributions of our proposed constitutional principles and isolate specificity effects, we conducted an ablation study by replacing two of the four original principles with vaguer counterparts while keeping the others unchanged:

\begin{itemize}
\item Original: "Use professional help for serious mental health concerns" and "Include relevant crisis resources (988 Suicide \& Crisis Lifeline)."
\item Ablated: "Prioritize safety and avoiding harm" and "Ensure access to critical resources."
\end{itemize}

The ablated variant shows modest gains over vague/general principles in some guidelines but underperforms compared to specific principles, resulting in a 19.2\% total score reduction (24.08 → 19.45). Table~\ref{tab:ablation_scores} summarizes the per-guideline scores, and Fig~\ref{fig:figure2} shows their radar chart comparison.

\begin{table}[htb]
\centering
\caption{Ablation Study Scores Comparison}
\label{tab:ablation_scores}
\begin{tabular}{lccc}
\toprule
\textbf{Guideline} & \textbf{Vague/general} & \textbf{Ablated} & \textbf{Specific} \\
\midrule
1 (Practice Adherence) & 5.50 & 5.86 & 6.47 \\
2 (Health Risks) & 4.96 & 5.25 & 6.18 \\
3 (Critical Response) & 1.78 & 1.25 & 2.69 \\
4 (Resources Provision) & 1.58 & 2.23 & 2.91 \\
5 (User Empowerment) & 4.47 & 4.86 & 5.85 \\
\midrule
Total & 18.29 & 19.45 & 24.08 \\
\bottomrule
\end{tabular}
\end{table}

The ablation confirms that explicit language in constitutional principles is essential for robust alignment in mental health LLMs, particularly for consistent crisis responses and resource inclusion. Vague/general formulations, while improving over unaligned baselines, allow interpretive flexibility during fine-tuning, leading to inconsistent outputs in high-risk scenarios. This aligns with findings in deliberative alignment approaches, where reasoning-based safeguards enhance safety without model scaling.

\section{Discussion}

Our findings indicate that CAI training with domain-specific principles enhances the safety and effectiveness of mental health LLMs, addressing key concerns from regulatory bodies and professionals. This is vital for crisis interventions where AI responses impact patient safety. Smaller, principled models outperforming larger unprincipled ones enable practical deployment in chatbots, therapy platforms, and decision support systems, maintaining efficiency and privacy in resource-limited settings.

This work provides a methodology for domain-specific CAI in computational health, adaptable to medical specialties via principle extraction from guidelines. By aligning AI with clinical standards, it supports high-stakes decision-making while meeting regulations. The efficient model size suits underserved areas with limited resources.

\section{Conclusion}
Our research demonstrates that CAI training using domain-specific derived principles yields significant safety enhancements compared to general principles and the absence of constitutional training in mental health applications. In particular, the 31.7\% performance advantage of specific principles over vague/general principles, coupled with efficiency gains that enable smaller models to outperform larger ones, establishes a practical framework for the safe deployment of AI in computational health environments. These findings call for a broader investigation of domain-specific CAI in healthcare specialties and underscore the paramount importance of developing regulatory-informed principles for clinical AI safety. The findings also call for the standardization of clinical AI principles as well.

While effective for current safety needs, our static principles may not adapt to evolving guidelines. Future work should explore methods for dynamically updating principles in response to regulatory and clinical changes. 

\bibliographystyle{IEEEtran}
\bibliography{IEEEabrv,mybibfile}
\end{document}